\documentclass[runningheads]{toolkit_includes/llncs}
\usepackage{graphicx}
\usepackage{subcaption}
\usepackage{tikz}
\usepackage{graphicx}
\usepackage{toolkit_includes/comment}
\usepackage{amsmath,amssymb} % define this before the line numbering.
\usepackage{color}
\usepackage{array}

\usepackage{hyperref}
\hypersetup{
    colorlinks=true,
    urlcolor=blue,
}

\newcommand*\blue{\color{blue}}
\newcommand*\red{\color{red}}

\makeatletter
\newcommand{\printfnsymbol}[1]{%
  \textsuperscript{\@fnsymbol{#1}}%
}
\makeatother

\begin{document}
% \renewcommand\thelinenumber{\color[rgb]{0.2,0.5,0.8}\normalfont\sffamily\scriptsize\arabic{linenumber}\color[rgb]{0,0,0}}
% \renewcommand\makeLineNumber {\hss\thelinenumber\ \hspace{6mm} \rlap{\hskip\textwidth\ \hspace{6.5mm}\thelinenumber}}
% \linenumbers
\pagestyle{headings}
\mainmatter
\def\ECCVSubNumber{32}  % Insert your submission number here

\title{Deep Atrous Guided Filter for Image Restoration in Under Display Cameras} % Replace with your title

%******************
% INITIAL SUBMISSION 
\begin{comment}
\titlerunning{ECCV-20 submission ID \ECCVSubNumber} 
\authorrunning{ECCV-20 submission ID \ECCVSubNumber} 
\author{Anonymous ECCV submission}
\institute{Paper ID \ECCVSubNumber}
\end{comment}

%******************
% CAMERA READY SUBMISSION
% TODO: Add author names post acceptance
% \begin{comment}
\titlerunning{Deep Atrous Guided Filter}
% If the paper title is too long for the running head, you can set
% an abbreviated paper title here
%
\author{Varun Sundar\thanks{Equal Contribution} \and
Sumanth Hegde\printfnsymbol{1} \and
Divya Kothandaraman \and
Kaushik Mitra}
\authorrunning{V. Sundar et al.}
% First names are abbreviated in the running head.
% If there are more than two authors, 'et al.' is used.

\institute{Indian Institute of Technology Madras \\
\email{\{varunsundar@smail, sumanth@smail, ee15b085@smail, kmitra@ee\}.iitm.ac.in}}
% \email{\{varunsundar,sumanth, divya\_kothandaraman\}@smail.iitm.ac.in, kmitra@ee.iitm.ac.in}}
% \end{comment}

%******************
\maketitle
\begin{abstract}
Under Display Cameras present a promising opportunity for phone manufacturers to achieve bezel-free displays by positioning the camera behind semi-transparent OLED screens. Unfortunately, such imaging systems suffer from severe image degradation due to light attenuation and diffraction effects. In this work, we present Deep Atrous Guided Filter (DAGF), a two-stage, end-to-end approach for image restoration in UDC systems. A Low-Resolution Network first restores image quality at low-resolution, which is subsequently used by the Guided Filter Network as a filtering input to produce a high-resolution output. Besides the initial downsampling, our low-resolution network uses multiple, parallel atrous convolutions to preserve spatial resolution and emulates multi-scale processing. Our approach's ability to directly train on megapixel images results in significant performance improvement. We additionally propose a simple simulation scheme to pre-train our model and boost performance. Our overall framework ranks 2nd and 5th in the RLQ-TOD'20 UDC Challenge for POLED and TOLED displays, respectively.
\end{abstract}

\section{Introduction}\label{introduction}

\begin{figure*}
    \centering
    \includegraphics[width=\textwidth]{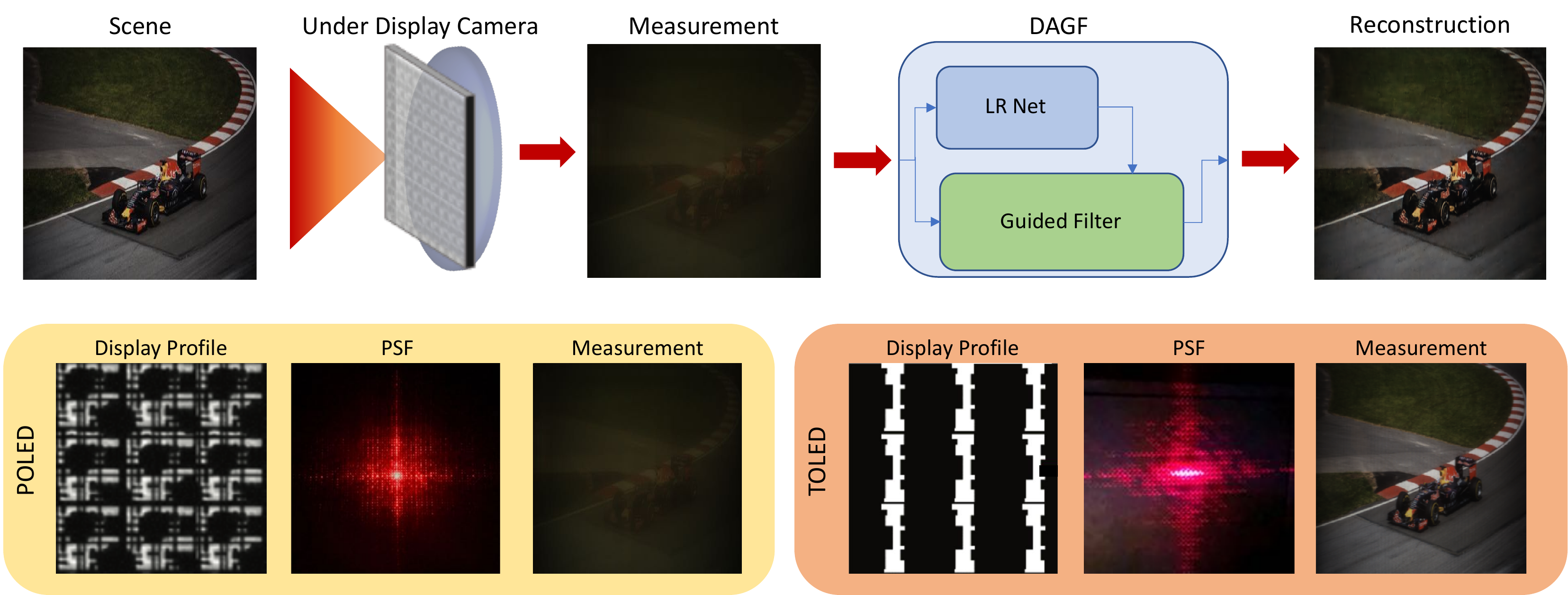}
    \caption{Under Display Cameras \cite{udc_zhou2020image} mount lenses behind semi-transparent OLED displays leading to image degradation. In this work, we introduce DAGF, which performs image restoration at megapixel resolution for both POLED and TOLED displays.}
    \label{fig:intro}
\end{figure*}

Under Display Cameras (UDC) promise greater flexibility to phone manufacturers by altering the traditional location of a smartphone's front camera. Such systems place the camera lens behind the display screen, making truly bezel-free screens possible and maximising screen-to-body ratio. Mounting the camera at the centre of the display also offers other advantages such as enhanced video call experience and is more relevant for larger displays found in laptops and TVs. However, image quality is greatly degraded in such a setup, despite the superior light efficiency of recent display technology such as OLED screens \cite{wenke2016organic}. As illustrated in Figure \ref{fig:intro}, UDC imaging systems suffer from a range of artefacts including colour degradation, noise amplification and low-light settings. This creates a need for restoration algorithms which can recover photorealistic scenes from UDC measurements.

Learning based methods, accentuated by deep learning, have achieved state-of-the-art performance on a variety of image restoration tasks including deblurring \cite{Brehm_2020_CVPR_Workshops,nah2020ntire,RUD_Sim_2019_CVPR_Workshops}, dehazing \cite{ancuti2020ntire,GCANet_chen2018gated,qin2020ffa}, denoising \cite{denoising_abdelhamed2020ntire,denoising_zhang2017beyond,denoising_zhang2018ffdnet}, deraining \cite{GCANet_chen2018gated,deraining_li2018recurrent} and image enhancement \cite{chen2017deeplab,HDRNet_gharbi2017deep}. However, deep learning techniques face two main drawbacks with regard to UDC imaging systems. First, such methods are do not scale computationally with input image resolution, and are typically run on much smaller patches. This is problematic for restoring severely degraded images such as UDC measurements, since small patches lack sufficient context. Second, common Convolutional Neural Networks (CNNs) employed in image restoration use multiple down-sampling operations to stack more layers and expand their receptive field without blowing up their memory footprint. Down-sampling leads to a loss of spatial information and affects performance in pixel-level dense prediction tasks such as image restoration \cite{Brehm_2020_CVPR_Workshops,multi_scale_Chen_2018_ECCV,learnt_downsampling_Marin_2019_ICCV}. An alternative is to simply omit such sub-sampling and resort to atrous (or dilated) convolutions. Owing to memory constraints, this is not feasible since we deal with high-resolution images in UDC systems.

To overcome these drawbacks, we propose a two-stage, end-to-end trainable approach utilizing atrous convolutions in conjunction with guided filtering. The first stage performs image restoration at low-resolution using multiple, parallel atrous convolutions. This allows us to maximally preserve spatial information without an exorbitant memory requirement. The guided filter then uses the low-resolution output as the filtering input to produce a high-resolution output via joint upsampling. Our approach makes it possible to directly train on high resolution images, and results in significant performance gains. Our contributions are as follows:

\begin{itemize}
\item We propose a novel image restoration approach for UDC systems utilizing atrous convolutions in conjunction with guided filters (Section \ref{our_method}).
\item We show that directly training on megapixel inputs allows our approach to significantly outperform existing methods (Section \ref{qual_quant_disc}). 
\item We propose a simple simulation scheme to pre-train our model and further boost performance (Section \ref{implementation}).
% \item We perform extensive ablative studies to determine the effectiveness of each component in our proposed approach (Section \ref{further_analysis}).
\end{itemize}

Our code and simulated data is publicly available at \href{https://varun19299.github.io/deep-atrous-guided-filter}{varun19299.github.io/deep-atrous-guided-filter/}.

% Note: trim first para a bit to accommodate "UDC Challenge line".

\section{Related Work}\label{related_work}

\textbf{Image restoration} encompasses tasks like image denoising, dehazing, deblurring and super resolution \cite{denoising_abdelhamed2020ntire,ancuti2020ntire,lugmayr2020ntire,nah2020ntire}. In recent years, deep learning has been the go-to tool in the field, with fully convolutional networks at the forefront of this success \cite{fully_conv_mao2016image,UNet_ronneberger2015u,fully_conv_tao2018scale,fully_conv_zhang2017learning}. Of these, residual dense connections \cite{RDN_zhang2020residual} exploiting hierarchical features has garnered interest with subsequent works in specific restoration tasks \cite{GCANet_chen2018gated,RDN_kim2019grdn,qin2020ffa,RDN_zhang2018density,RDN_zhang2020residual}. Another class of techniques use a GAN \cite{GANs_NIPS2014_5423} based setting. Methods like \cite{jiang2019enlightengan,kupyn2019deblurgan} fall in this category. Finally, there exist recent work exploiting CNNs as an effective image prior \cite{lehtinen2018noise2noise,ulyanov2018deep,zhang2017learning}. However, the above-mentioned methods operate on small patches of the input image and do not scale to larger input dimensions. 

\noindent \textbf{Joint upsampling} seeks to generate a high-resolution output, given a low-resolution input and a high-resolution guidance map. Joint Bilateral Upsampling \cite{joint_bilateral_upsampling} uses a bilateral filter towards this goal, obtaining a piecewise-smooth high-resolution output, but at a large computational cost. Bilateral Grid Upsampling \cite{bilateral_guided_upsampling} greatly alleviates this cost by fitting a grid of local affine models on low-resolution input-output pairs, which is then re-used at high resolution. Deep Bilateral Learning \cite{HDRNet_gharbi2017deep} integrates bilateral filters in an end-to-end framework, with local affine grids that can be learnt for a particular task. 

\noindent \textbf{Guided filters} \cite{Guided_filter_He} serve as an alternative to Joint Bilateral Upsampling, with superior edge-preserving properties at a lower computational cost. Deep Guided Filtering \cite{DGF_wu2017fast} integrates this with fully convolutional networks and demonstrates it for common image processing tasks, with recent interest in the hyperspectral \cite{guo2018guided}, remote \cite{xu2018building} and medical imaging \cite{gong2017boosting}. Guided filters have been mainly explored in the context of accelerating image processing operators. In our work, we present a different application of image restoration.  

\noindent \textbf{Atrous or dilated convolutions} incorporate a larger receptive field without an increase in the number of parameters or losing spatial resolution. Yu \textit{et al.} \cite{yu2017dilated} proposed a residual network using dilated convolutions. Dilated convolutions have found success in semantic segmentation \cite{chen2017deeplab,zhao2017pyramid}, dehazing \cite{GCANet_chen2018gated,qin2020ffa} and deblurring \cite{Brehm_2020_CVPR_Workshops} tasks as well as general image processing operations \cite{CAN_Chen_2017_ICCV}. However, a major challenge in atrous networks is keeping memory consumption in check. Multi-scale fusion via pyramid pooling or encoder-decoder networks \cite{multi_scale_chen2017rethinking,multi_scale_Chen_2018_ECCV,multi_scale_Lin_2018_ECCV,multi_scale_Sarker_2018_ECCV_Workshops,zhang2017learning} can offload intensive computation to lower scales, but can lead to missing fine details. Instead, we include channel and pixel attention \cite{qin2020ffa} to have a flexible receptive field at each stage while better tending to severely degraded regions.

Compared to prior work, \textbf{our main novelty} lies in directly training on megapixel images by  incorporating multiple, parallel smoothed atrous convolutions in a guided filter framework. This adapts the proposed framework in Wu \textit{et al.} \cite{DGF_wu2017fast}- primarily developed for image processing tasks- to handle the challenging scenario of image restoration for Under Display Cameras.

\section{Deep Atrous Guided Filter}\label{our_method}

\begin{figure*}[!t]
    \centering
    \includegraphics[width=0.95\textwidth]{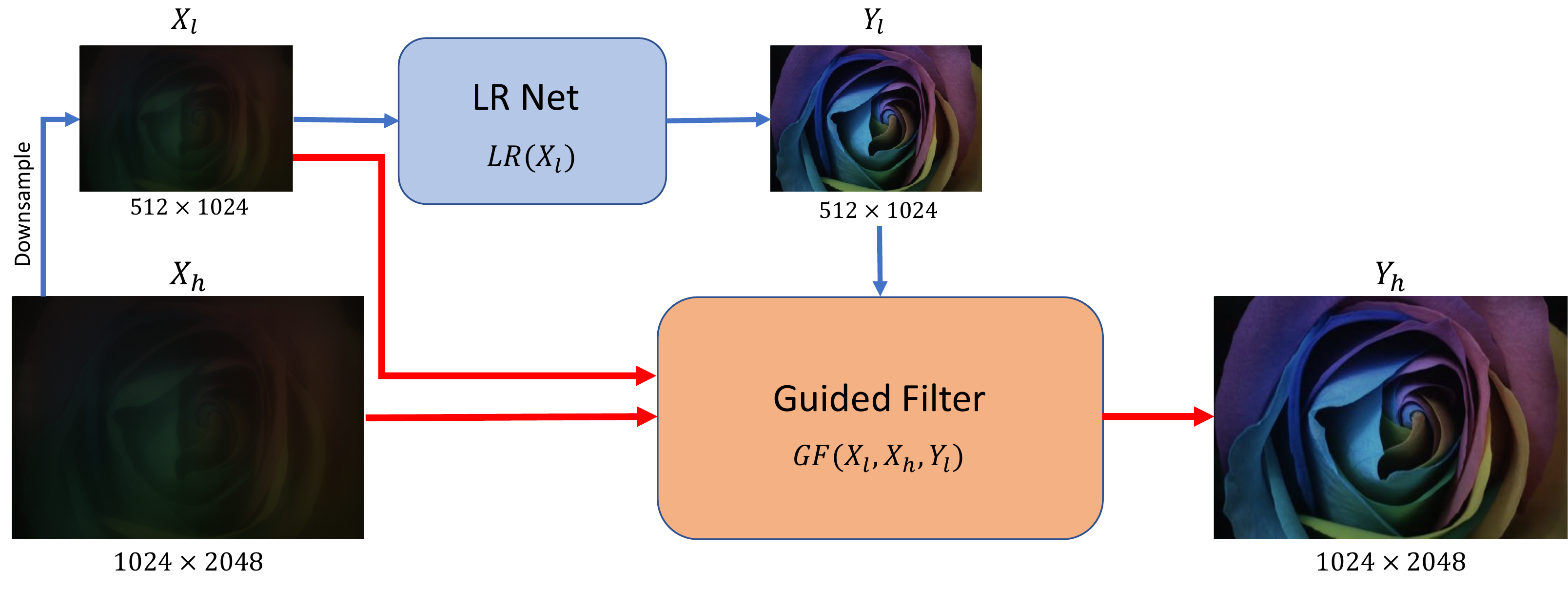}
    \caption{\textbf{Framework overview of DAGF.} Our architecture seeks to operate directly on megapixel images by performing joint upsampling. A low resolution network (LRNet) restores a downsampled version $X_l$ of input $X_h$ to produce $Y_l$. The guided filter then uses this to yield the final high-resolution output $Y_h$.}
    \label{fig:our_method}
\end{figure*}

To address the challenges posed by Under Display Cameras, we employ a learning based approach that directly trains on megapixel images. We argue that since UDC measurements are severely degraded, it is imperative to train models with large receptive fields on high-resolution images \cite{large_receptive_Kim_2016_CVPR,RUD_Sim_2019_CVPR_Workshops,large_receptive_Simonyan15}. 

Our approach, \textbf{Deep Atrous Guided Filter Network (DAGF)}, consists of two stages: (a) Low Resolution Network (LRNet), which performs image restoration at a lower resolution, and (b) Guided Filter Network, which uses the restored low-resolution output and the high-resolution input to produce a high-resolution output. Our guided filter network, trained end-to-end with LRNet, restores content using the low-resolution output while preserving finer detail from the original input.

We design our approach to perform image restoration for two types of OLED displays: Pentile OLED (POLED) and  Transparent OLED (TOLED). As seen in Figure \ref{fig:intro}, TOLED has a stripe pixel layout, while POLED has a pentile pixel layout with a much lower light transmittance. Consequently, TOLED results in a blurry image, while POLED results in a low-light, colour-distorted image.

\subsection{LR Network}\label{lr_net}

LRNet comprises of three key components: i) PixelShuffle \cite{pixelshuffle_shi2016real} ii) atrous residual blocks, and iii) a gated attention mechanism \cite{GCANet_chen2018gated,Tai-MemNet-2017}. We first use PixelShuffle \cite{pixelshuffle_shi2016real} to lower input spatial dimensions while expanding channel dimensions. This affords us a greater receptive field at a marginal memory footprint \cite{uses_pixelshuffle_gu2019self,edsr_Lim_2017_CVPR_Workshops}. We further encode the input into feature maps via successive atrous residual blocks and then aggregate contextual information at multiple levels by using a gated attention mechanism. We now describe each component of LRNet. \\

\begin{figure*}[t]
    \centering
    \includegraphics[width=\textwidth]{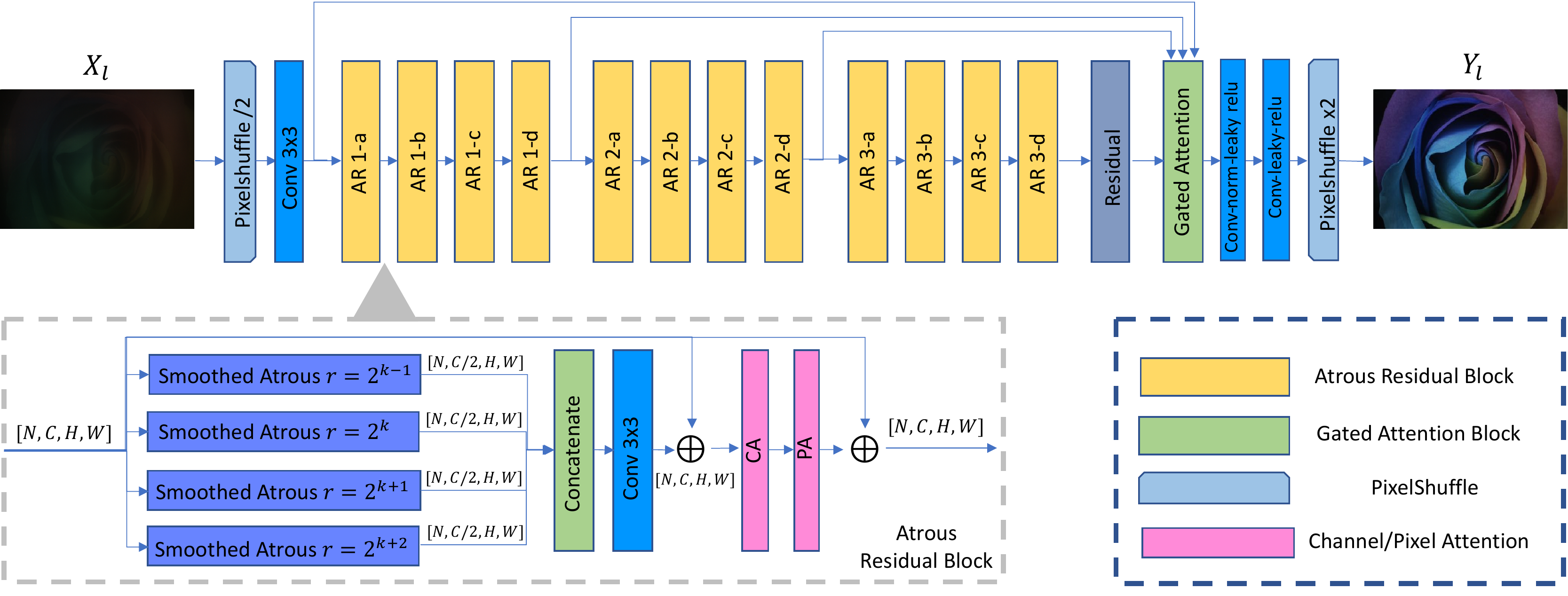}
    \caption{\textbf{Overview of LRNet.} LRNet operates on a low-resolution version $X_l$ of original input $X_h$. The input image $X_l$ is downsampled via pixelshuffle, encoded via many atrous residual blocks and finally a gated attention mechanism aggregates contextual information to produce low-resolution output $Y_l$.}
    \label{fig:lr_network}
\end{figure*}

\noindent\textbf{Smoothed Atrous Convolutions.} Unlike common fully convolutional networks employed in image restoration \cite{uses_pixelshuffle_gu2019self,edsr_Lim_2017_CVPR_Workshops,PANet_mei2020pyramid}, which use multiple downsampling blocks, we opt to use atrous (or dilated) convolutions \cite{dialated_Yu:2016:MCA} instead. This allows us expand the network's receptive field without loss in spatial resolution, which is beneficial for preserving fine detail in dense prediction tasks.
 
Atrous convolutions, however, lead to gridding artefacts in their outputs \cite{gridding_wacv_2018,gridding_wacv_2018_wang,griddingwang2018smoothed}. To alleviate this, we insert a convolution layer before each dilated convolution, implemented via shared separable kernels for computational and parameter efficiency \cite{griddingwang2018smoothed}. Concretely, for a input feature map $F^\text{in}$ with $C$ channels, the smoothed atrous convolution layer produces output feature map $F^\text{out}$ with $C$ channels as follows:

\begin{equation}
    F^\text{out}_i =  \sum_{j \in [C]}\big( (F^\text{in}_j * K^{\text{sep}} + b_i)*_r K_{ij} \big)
\end{equation}

\noindent where $F^\text{out}_i$ is the $i$\textsuperscript{th} output channel, $b_i$ is a scalar bias, $*$ is a 2D convolution and $*_r$ is a dilated convolution with dilation $r$. $K_{ij}$ is a $3 \times 3$ convolution kernel and $K^\text{sep}$ is the shared separable convolution kernel, shared among all input feature channels. For dilation rate $r$, we use a shared separable kernel of size $2r - 1$.

We also add adaptive normalization \cite{CAN_Chen_2017_ICCV} and leaky rectified linear unit (LReLU) after the smoothed atrous convolution. LReLU may be represented as: $\Phi(x)=\max(\alpha x,x)$, where we set $\alpha=0.2$. Adaptive Normalization combines any normalization layer and the identity mapping as follows:

\begin{equation}
    AN(F^\text{in}) = \lambda F^\text{in} + \mu N(F^\text{in})
\end{equation}

\noindent where $F^\text{in}$ is the input feature map, $\lambda,\mu \in \mathbb{R}$ and $N(.)$ is any normalization layer such as batch-norm \cite{ioffe2015batch} or instance-norm \cite{instance_ulyanov2017improved}. We use instance-norm in our adaptive normalization layers. In our ablative studies (Section \ref{ablative_studies}), we show that our adaptive normalization layer results in improved performance. \\

\noindent\textbf{Atrous Residual blocks.} As depicted in Fig. \ref{fig:lr_network}, we propose to use multiple, parallel, smoothed atrous convolutions with various dilation rates in our residual blocks, following its recent success in image deblurring \cite{Brehm_2020_CVPR_Workshops}. For atrous residual block AR-k, belonging to the $k$th group, we use four smoothed atrous convolutions with dilation rates $\{2^{k-1}, 2^k, 2^{k+1}, 2^{k+2}\}$. Each convolution outputs a feature map with $C/2$ channels, which we concatenate to obtain $2C$ channels. These are subsequently reduced to $C$ channels via a $1 \times 1$ convolution. Our atrous residual blocks also utilize channel and pixel attention mechanisms, which are described below. \\

\noindent\textbf{Channel Attention.} We use the channel attention block proposed by Qin \textit{et al.} \cite{qin2020ffa}. Specifically for a feature map $F^\text{in}$ of dimensions $C \times H \times W$, we obtain channel-wise weights by performing global average pooling (GAP) and further encode it via two $1 \times 1$ conv layers. We multiply $F^\text{in}$ with these channel weights $CA$ to yield output $F^\text{out}$:

\begin{equation}\label{eq:global_avg_pooling}
        GAP_i = \frac{1}{HW} \sum_{u \in [H], v \in [W]} F^\text{in}_i(u,v)
\end{equation}
\begin{equation}\label{eq:CA_conv}
        CA_i = \sigma \big( \sum_{j \in [C]}\Phi(\sum_{k \in [C/8]} GAP_k * K_{jk} + b_j) * K^\prime_{ij} + b_i \big)
\end{equation}
\begin{equation}\label{eq:CA_mul}
F^\text{out}_i = CA_i \odot F^\text{in}_i
\end{equation}

\noindent where, $\sigma$ is the sigmoid activation, $\Phi$ is LReLU described earlier and $\odot$ is element-wise multiplication. \\

\noindent\textbf{Pixel Attention.} To account for uneven context distribution across pixels, we use a pixel attention module \cite{qin2020ffa} that multiplies the input feature map $F^\text{in}$  of shape $C \times H \times W$ with an attention map of shape $1 \times H \times W$ varying across pixels, but constant across channels. We obtain the pixel attention map $PA$ by using two $1 \times 1$ conv layers:

\begin{equation}\label{eq:PA_conv}
        PA = \sigma \big( \sum_{j \in [C/8]}\Phi(\sum_{k \in [C]}F^\text{in}_k * K_{jk} + b_j) * K^\prime_{j} + b \big)
\end{equation}
\begin{equation}\label{eq:PA_mul}
F^\text{out}_i = PA \odot F^\text{in}_i
\end{equation}

\noindent\textbf{Gated Attention.} We utilise a gated attention mechanism \cite{GCANet_chen2018gated,Tai-MemNet-2017} to aggregate information across several atrous residual blocks. Fusing features from different levels is beneficial for both low-level and high-level tasks \cite{FPN_Lin_2017_CVPR,dense_pyramid_Zhang_2018_CVPR,zhao2017pyramid}. We extract feature maps before the first atrous residual block ($F^0$), and right after each atrous residual group ($F^1,...,F^k$). For $k$ atrous groups, we concatenate these $k+1$ feature maps and output $k+1$ masks, using $\mathcal{G}$, a $3 \times 3$ conv layer:

\begin{equation}\label{eq:gated_masks}
(\mathcal{M}^0,\mathcal{M}^1,..., \mathcal{M}^k) = \mathcal{G}(F^0,F^1,...,F^k) 
\end{equation}
\begin{equation}\label{eq:gated_concat}
F^\text{out} = \mathcal{M}^0 \odot F^0 + \sum_{l\in[k]} \mathcal{M}^l \odot F^l
\end{equation}

\subsection{Guided Filter Network}

\begin{figure*}[t]
    \centering
    \includegraphics[width=0.95\textwidth]{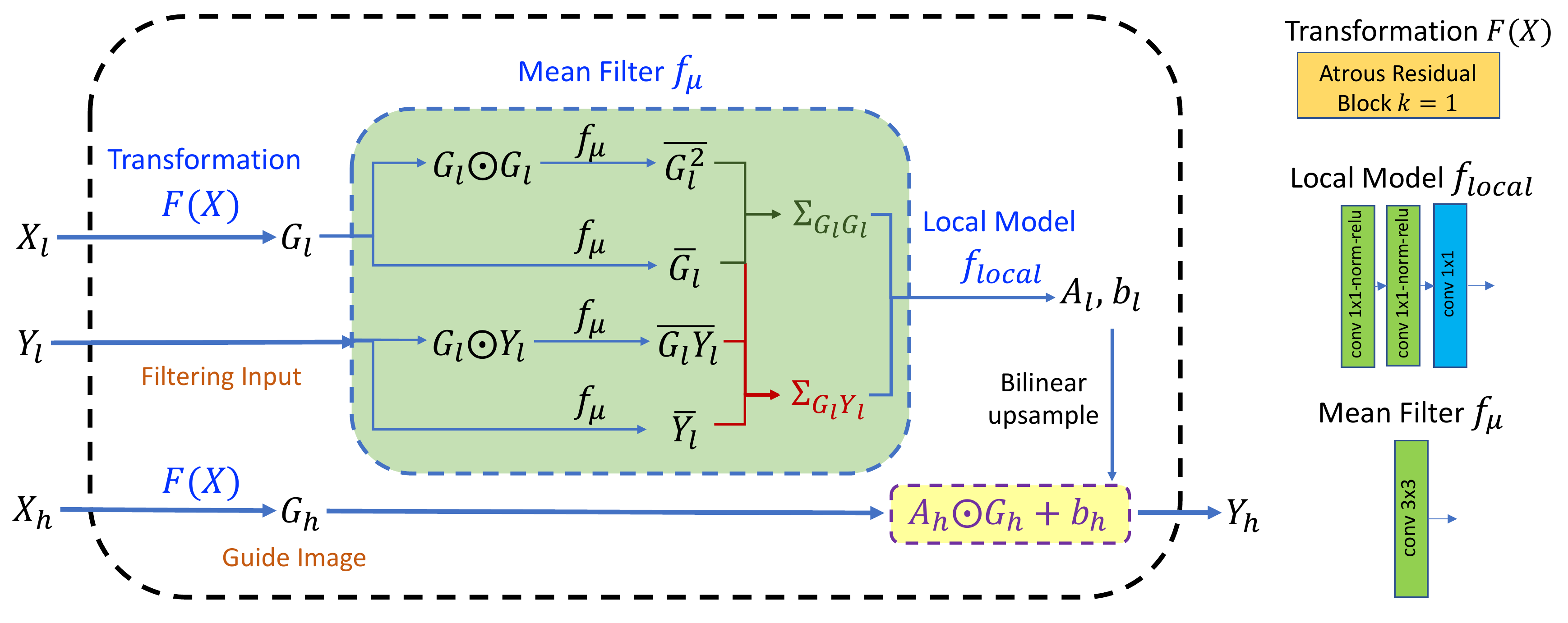}
    \caption{\textbf{Computational Graph of Guided Filter Stage.} The guided filter first transforms the high-resolution input $X_h$ to guide image $G_h$, and then yields the final output $Y_h$ via joint upsampling. Our guided filter network is differentiable and end-to-end trainable \cite{DGF_wu2017fast}.
    % The guided filter accepts a high-resolution input $X_h$, low-resolution input $X_l$ and low-resolution output $Y_l$. Our guided filter layer is differentiable and end-to-end trainable \cite{DGF_wu2017fast}. 
    % {\red We implement its learnable components via convolutional blocks; specifically the guidance map $F(X)$ as a \textbf{atrous residual block}, mean filter $f_\mu$ as a $3 \times 3$ convolution and local model $f_\text{local}$ as a 3 layer conv block.} 
    }
    \label{fig:guided_filter}
\end{figure*}

Given a high-resolution input $X_h$, low-resolution input $X_l$ and low-resolution output $Y_l$, we seek to produce a high-resolution output $Y_h$, which is perceptually similar to $Y_l$ while preserving fine detail from $X_h$. We adopt the guided filter proposed by He \textit{et al.} \cite{Guided_filter_He,he2015fast} and use it in an end-to-end trainable fashion \cite{DGF_wu2017fast}. As illustrated in Figure \ref{fig:guided_filter}, the guided filter formulates $Y_h$ as:

\begin{equation}
    Y_h = A_h \odot G_h + b_h
\end{equation}

\noindent where $G_h = F(X_h)$ is a transformed version of input $X_h$. We bilinear upsample $A_h$ and $b_h$ from low-resolution counterparts $A_l$ and $b_l$, such that:

\begin{equation}\label{eq:local_linear_model}
\begin{split}
    \overline{Y}_l = A_l \odot \overline{G}_l + b_l \\
\end{split}
\end{equation}

\noindent where $\overline{G}_l,\overline{Y}_l$ are mean filtered versions of $G_l,Y_l$, ie., $\overline{G}_l = f_\mu(G_l)$ and $\overline{Y}_l = f_\mu(Y_l)$. Compared to Wu \textit{et al.} \cite{DGF_wu2017fast}, we implement $f_\mu$ by a $3 \times 3$ convolution (instead of a box-filter). Instead of directly inverting Equation \ref{eq:local_linear_model}, we obtain its solution using $f_\text{local}$, implemented by a 3 layer, $1 \times 1$ convolutional block:

\begin{equation}\label{eq:solve_local_linear_model}
\begin{split}
    A_l = f_\text{local}(\Sigma_{G_lY_l}, \Sigma_{G_lG_l}), b_l = \overline{Y}_l - A_l \odot \overline{G}_l
\end{split}
\end{equation}

\noindent where covariances are determined as, $\Sigma_{G_lY_l} = \overline{G_lY_l} - \overline{G}_l\overline{Y}_l$, etc. Finally, we use our atrous residual block to implement the transformation function $F(X)$, and show that it confers substantial performance gains (Section \ref{effect_guided_filter}). Overall, our guided filter consists of three trainable components, viz. $F$, $f_\mu$ and $f_\text{local}$.

\subsection{Loss Function}

\noindent\textbf{L1 Loss.} We employ Mean Absolute Error (or L1 loss) as our objective function. We empirically justify our choice L1 loss over other loss formulations (including MS-SSIM \cite{MS-SSIM_zhao2016loss}, perceptual \cite{johnson2016perceptual} and adversarial \cite{GANs_NIPS2014_5423} losses) in Section \ref{ablative_studies}.

\section{Experiments and Analysis}\label{experiments}

\subsection{Dataset} 

Our network is trained on the POLED and TOLED datasets \cite{udc_zhou2020image} provided by the UDC 2020 Image Restoration Challenge. Both datasets comprise of 300 images of size $1024 \times 2048$, where 240 images are used for training, 30 for validation and 30 for testing in each track. We do not have access to any specific information of the forward model (such as the PSF or display profile), precluding usage of non-blind image restoration methods such as Wiener Filter \cite{wiener_filter_orieux2010bayesian}.

\subsection{Implementation Details}\label{implementation}

\noindent\textbf{Model Architecture.} LRNet comprises of 3 atrous residual groups, with 4 blocks each. The intermediate channel size in LRNet is set to 48. The training data is augmented with random horizontal flips, vertical flips and $180^{\circ}$ rotations. All images are normalized to a range between -1 and 1. The AdamW \cite{loshchilov2018decoupled} optimizer, with initial learning rate $\eta=0.0003$, $\beta_1=0.9$ and $\beta_2=0.999$ is used. The learning rate is varied with epochs as per the cosine annealing scheduler with warm restarts \cite{loshchilov-ICLR17SGDR}. We perform the first warm restart after 64 epochs, post which we double the duration of each annealing cycle. The models are trained using PyTorch \cite{Pytorch_NEURIPS2019_9015} on 4 NVIDIA 1080Ti GPUs with a minibatch size of 4, for 960 epochs each. \\

\noindent\textbf{Pre-training Strategy.} To aid in faster convergence and boost performance, we pre-train our model with simulated data. The UDC dataset is created using monitor acquisition \cite{udc_zhou2020image}, where images from the DIV2K dataset \cite{DIV2K_Agustsson_2017_CVPR_Workshops} are displayed on a LCD monitor and captured by a camera mounted behind either glass (considered ground-truth) or POLED/TOLED panels (low-quality images). To simulate data, we need to transform clean images from the DIV2K dataset to various display measurements (POLED, TOLED or glass).

\begin{figure*}[!t]
    \centering
    \includegraphics[width=0.95\textwidth]{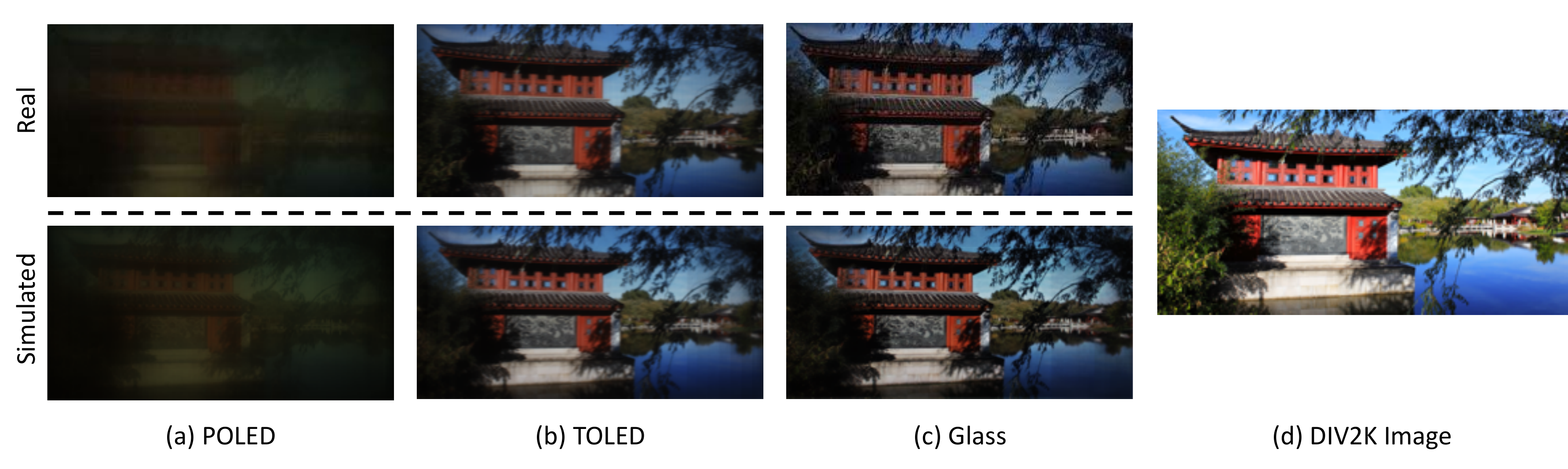}
    \caption{\textbf{Pre-training using simulated data enhances performance.} We transform 800 DIV2K \cite{DIV2K_Agustsson_2017_CVPR_Workshops} images via a simulation network to various display measurements (Glass, TOLED and POLED). To train our simulation network, we use the misalignment tolerant CoBi \cite{CoBi_Zhang_2019_CVPR} loss.
    }
    \label{fig:simulation}
\end{figure*}

Using Fresnel Propagation to simulate data with either the display profile or calibrated PSF can be inaccurate \cite{udc_zhou2020image}. Instead, a shallow variant of our model is trained to transform 800 images from the DIV2K dataset to each measurement. Since DIV2K images do not align with display measurements, we leverage the Contextual Bilateral (CoBi) Loss \cite{CoBi_Zhang_2019_CVPR}, which can handle moderately misaligned image pairs. For two images $P$ and $Q$, with $\{\vec{p}_{ij}\}$ and $\{\vec{q}_{ij}\}$ ($i \in [H],j \in [W]$) representing them as a grid of RGB intensities, CoBi loss can be written as:

\begin{equation}
    \text{CoBi}(P,Q) = \frac{1}{HW} \sum_{i,j} \min_{k,l} \left[ \mathbb{D}(\vec{p}_{ij},\vec{q}_{kl}) + \gamma \left( (i-k)^2 + (j-l)^2 \right) \right]
\end{equation}

\noindent where $\mathbb{D}$ is any distance metric (we use cosine distance). $\gamma$ allows CoBi to be flexible to image-pair misalignment. As seen in Figure \ref{fig:simulation}, our simulated measurements closely match real measurements. Such an initialisation procedure gives our model (DAGF-PreTr) around 0.3 to 0.5 dB improvement in PSNR (Table \ref{tab:main_metrics}). More simulation results can be found in the supplementary material.

\subsection{Quantitative and Qualitative Analysis}\label{qual_quant_disc}

\noindent\textbf{Baseline Methods.} Our method is compared against four image restoration methods: DGF \cite{DGF_wu2017fast}, PANet \cite{PANet_mei2020pyramid}, UNet \cite{UNet_ronneberger2015u}  and FFA-Net \cite{qin2020ffa}. DGF utilises a trainable guided filter for image transformation. For DGF, we use 9 layers in the CAN \cite{CAN_Chen_2017_ICCV} backbone (instead of 5) for better performance. UNet is a popular architecture in image restoration. A variant of UNet with a double encoder \cite{udc_zhou2020image} and 64 intermediate channels in the first block is used. PANet and FFA-Net are specifically designed architectures for image denoising and dehazing, respectively. Small patch sizes often provide little information for faithful restoration. Hence, to make a fair comparison, a much larger patch size of $96 \times 96$ for PANet (compared to $48 \times 48$ in Mei \textit{et al.} \cite{PANet_mei2020pyramid}), $256 \times 512$ for UNet ($256 \times 256$ in Zhou \textit{et al.} \cite{udc_zhou2020image}) and $256 \times 512$ for FFA-Net ($240 \times 240$ in Qin \textit{et al.} \cite{qin2020ffa}) is used. \\

% \footnote{All methods are implemented using publicly available code.} \\

\noindent\textbf{Quantitative and Qualitative Discussion.} All our methods are evaluated on PSNR, SSIM and the recently proposed LPIPS \cite{LPIPS_zhang2018unreasonable} metrics. Higher PSNR and SSIM score indicate better performance, while lower LPIPS indicates better perceptual quality. As seen in Table \ref{tab:main_metrics}, our approach (DAGF) significantly outperforms the baselines, with an improvement of $3.2$ dB and $0.5$ dB over the closest baseline on POLED and TOLED measurements, respectively. 

\begin{figure*}[!th]
\centering

\begin{subfigure}{0.95\textwidth}
\centering
\includegraphics[width=\textwidth]{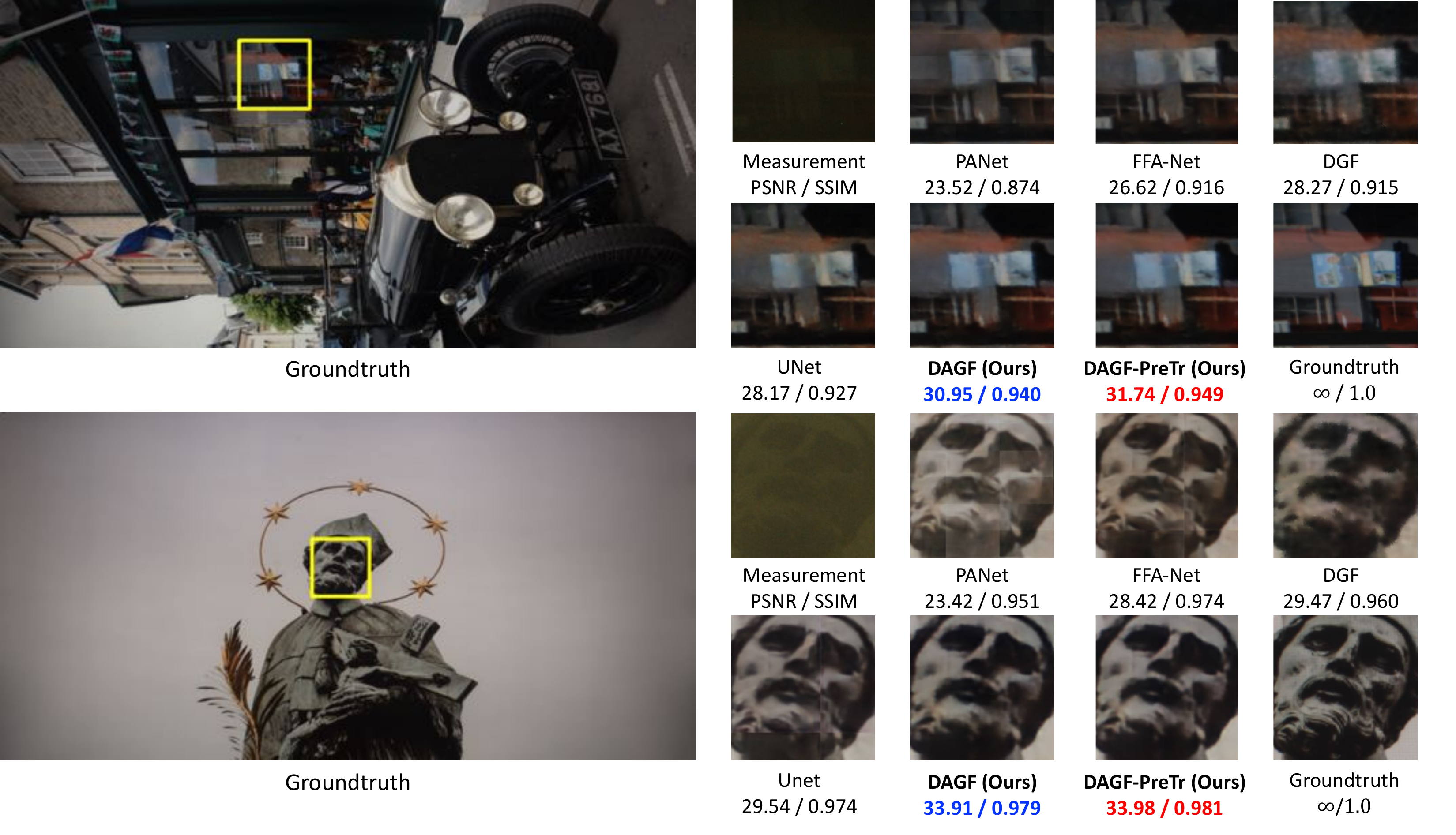}
\caption{\textsc{POLED} dataset.}
\label{fig:poled_comp}
\end{subfigure}

\begin{subfigure}{0.95\textwidth}
\centering 
\includegraphics[width=\textwidth]{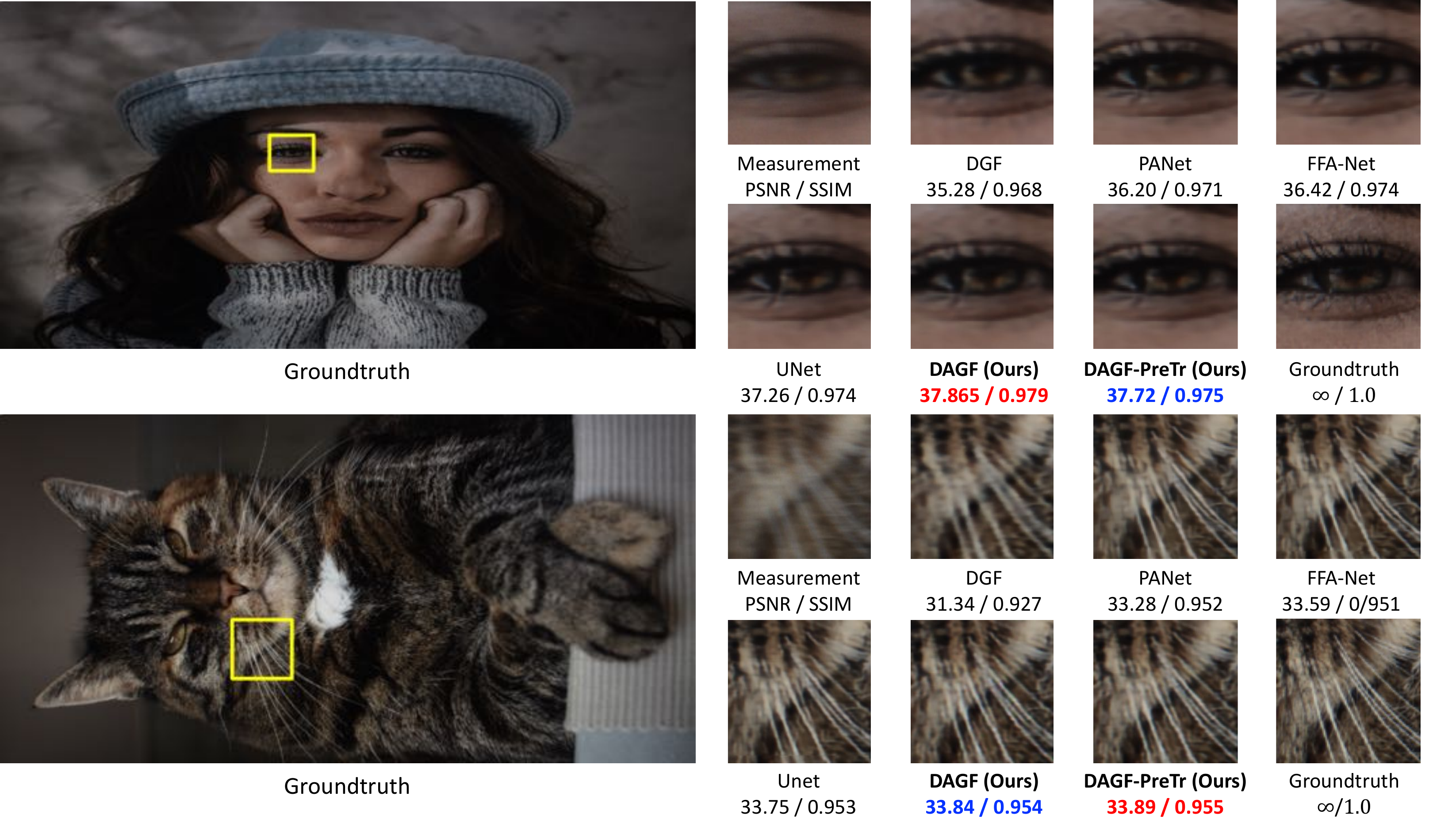}
\caption{\textsc{TOLED} dataset.}
\label{fig:toled_comp}
\end{subfigure}

\caption{\textbf{Qualitative results.} DAGF is considerably superior to patch based restoration methods  \cite{PANet_mei2020pyramid,qin2020ffa,UNet_ronneberger2015u}, more evident on the severely degraded POLED measurements. Metrics evaluated on entire image. Zoom in to see details.}
\label{fig:visual_comp}
\end{figure*}

Our approach's ability to directly train on megapixel images and hence aggregate contextual information over large receptive fields leads to a significant improvement. This is more evident on the POLED dataset, where patch based methods such as PANet, UNet and FFA-Net lack sufficient context despite using larger patch-sizes. With the exception of DGF, our approach also uses much lesser parameters. Visual comparisons in Figure \ref{fig:visual_comp} are consistent with our quantitative results. Our approach closely resembles groundtruth, having lesser artefacts and noise. Notably, in Figure \ref{fig:poled_comp}, we can observe line artefacts in patch based methods (further detailed in Section \ref{effect_guided_filter}). \\

\begin{table}[!t]
    \centering
    \caption{\textbf{Quantitative comparison.} By directly training on megapixel images, our approach, DAGF significantly outperforms baselines. To further boost performance, we pre-train on simulated data (DAGF-PreTr). {\red Red} indicates the best and {\blue Blue} the second best in the chosen metric (on validation set).}
    
    \resizebox{0.95\textwidth}{!}{\begin{tabular}{ c||c|c c c|c c c }
    \hline
    \textbf{Method}&\textbf{\#Params $\downarrow$}& 
    \multicolumn{3}{c|}{\textbf{POLED}} & \multicolumn{3}{c}{\textbf{TOLED}}\\
    \hline
    {}  & {} & 
    PSNR $\uparrow$ & SSIM $\uparrow$ & LPIPS $\downarrow$ & 
    PSNR $\uparrow$ & SSIM $\uparrow$ & LPIPS $\downarrow$ \\
    \hline
    PANet \cite{PANet_mei2020pyramid} & 6.0M & 
    {26.22} & {0.908} & {0.308} & 
    {35.712} & {0.972} & {0.147} \\
    \hline
    FFA-Net \cite{qin2020ffa} & 1.6M & 
    {29.02} & {0.936} & {0.256} & 
    {36.33} & {0.975} & {\red 0.126} \\
    \hline
    DGF \cite{DGF_wu2017fast} & {\red 0.4M} & 
    29.93 & 0.931 & 0.362 & 
    34.43 & 0.956 & 0.220 \\
    \hline
    Unet \cite{UNet_ronneberger2015u} & 8.9M & 
    {29.98} & {0.932} & {0.251} & 
    {36.73} & {0.971} & {0.143} \\
    \hline
    \textbf{DAGF (Ours)}   & {\blue 1.1M} &
    {\blue 33.29} & {\blue 0.952} & {\blue 0.236} &
    {\blue 37.27} & {\blue 0.973} & 0.141\\
    \hline
    \textbf{DAGF-PreTr (Ours)} & {\blue 1.1M} &
    {\red 33.79} & {\red 0.958} & {\red 0.225} &
    {\red 37.57} & {\red 0.973} & {\blue 0.140} \\
    \hline
    \end{tabular}
    }
    \label{tab:main_metrics}
\end{table}

% \noindent\textbf{Computational Analysis.} Table \ref{tab:main_metrics} also features the memory consumption, parameter count and runtime (inference time) of each method. Memory consumption and runtime are both evaluated only for the forward pass on a NVIDIA GTX 1080Ti GPU. For patch-based methods, we try stacking all patches along the batch dimension, if memory permits. 

% DGF, which also uses a trainable guided filter, has the least computational footprint, but provides inferior performance. In comparison, our method provides superior performance at moderate parameter count and memory requirement, but with slightly longer inference time. The runtime of patch-based methods is dependent on fitting all patches within the given memory constraints, hence having either high memory requirement or long inference time. Although not compared here, it is possible to trade-off performance for memory and runtime in our method by using a larger downscaling factor for LRNet. \\

% \noindent\textbf{Geometric Ensembling.} Incorporating geometric self-ensemble in our pipeline \cite{RUD_Sim_2019_CVPR_Workshops,7_ways_Timofte_2016_CVPR} further boosts performance by augmenting input images with rotated and flipped versions. All transformations are fed into the network and their output is inverse transformed before averaging. We refer to geometric ensembled models by adding a `+' postfix to the method name, i.e. DAGF-PreTr+ (Table \ref{tab:udc_challenge}). 

\noindent\textbf{Challenge Results.} This work is initially proposed for participating in the UDC 2020 Image Restoration Challenge \cite{udc_challenge_zhou2020}. For the challenge submission, geometric self-ensembling \cite{RUD_Sim_2019_CVPR_Workshops,7_ways_Timofte_2016_CVPR} is incorporated in DAGF-PreTr to boost performance, denoted in Table \ref{tab:udc_challenge} as DAGF-PreTr+. Self-ensembling involves feeding various rotated and flipped versions of the input image to the network, and performing corresponding inverse transforms before averaging their outputs. 
% This strategy is advantageous over other ensembling techniques since it does not require training multiple models, while offering comparable performance gains \cite{edsr_Lim_2017_CVPR_Workshops}.

% We refer to the geometric ensembled model by adding a `+' postfix to the method name, i.e., DAGF-PreTr+ (Table \ref{tab:udc_challenge}).

Quantitatively, our method ranks 2nd and 5th on the POLED and TOLED tracks, respectively (Table \ref{tab:udc_challenge}), proving that DAGF is effective at image restoration, especially in the severe degradation setting of POLED. While our approach is competitive on both tracks, there is scope to better adapt our model to moderate image degradation scenarios such as TOLED measurements.

\begin{table}[!t]
\centering
\caption{\textbf{Comparison on UDC2020 Image Restoration Challenge.} {\red Red} indicates the best performance and {\blue Blue} the second best (on challenge test set).}

    \resizebox{0.95\textwidth}{!}{\begin{tabular}{| c | c c || c | c c |}
    \hline
    \multicolumn{3}{|c||}{\textbf{POLED}}  & 
    \multicolumn{3}{c|}{\textbf{TOLED}}  \\
    \hline
    Method  & 
    PSNR $\uparrow$ & SSIM $\uparrow$ & 
    Method  & 
    PSNR $\uparrow$ & SSIM $\uparrow$ \\
    \hline
    First Method &
    {\red 32.99} & {\red 0.957} &
    First Method &
    {\red 38.23} & {\red 0.980} \\
    \hline
    \multicolumn{1}{|m{4cm}|}{\centering \textbf{DAGF-PreTr+ (Ours)}} &
    {\blue 32.29} & {\blue 0.951} &
    Second Method &
    {\blue 38.18} & {\blue 0.980} \\
    \hline
    Third Method &
    {31.39} & {0.950} &
    Third Method &
    {38.13} & {0.980} \\
    \hline
    Fourth Method &
    {30.89} & {0.947} &
    Fourth Method &
    {37.83} & {0.978} \\
    \hline
    Fifth Method &
    {29.38} & {0.925} &
    \multicolumn{1}{m{4cm}|}{\centering \textbf{DAGF-PreTr+ (Ours)}} &
    {36.91} & {0.973} \\
    \hline
    \end{tabular}}

\label{tab:udc_challenge}
\end{table}
\section{Further Analysis}\label{further_analysis}

To understand the role played by various components in DAGF, extensive ablative studies have been conducted. These experiments have been performed on downsized measurements, i.e., $512 \times 1024$, in order to reduce training time.

\subsection{Effect of Guided Filter}\label{effect_guided_filter}

\begin{figure}[!t]
  \centering
  \begin{minipage}[b]{0.48\textwidth}
    \centering
    \includegraphics[width=0.8\textwidth]{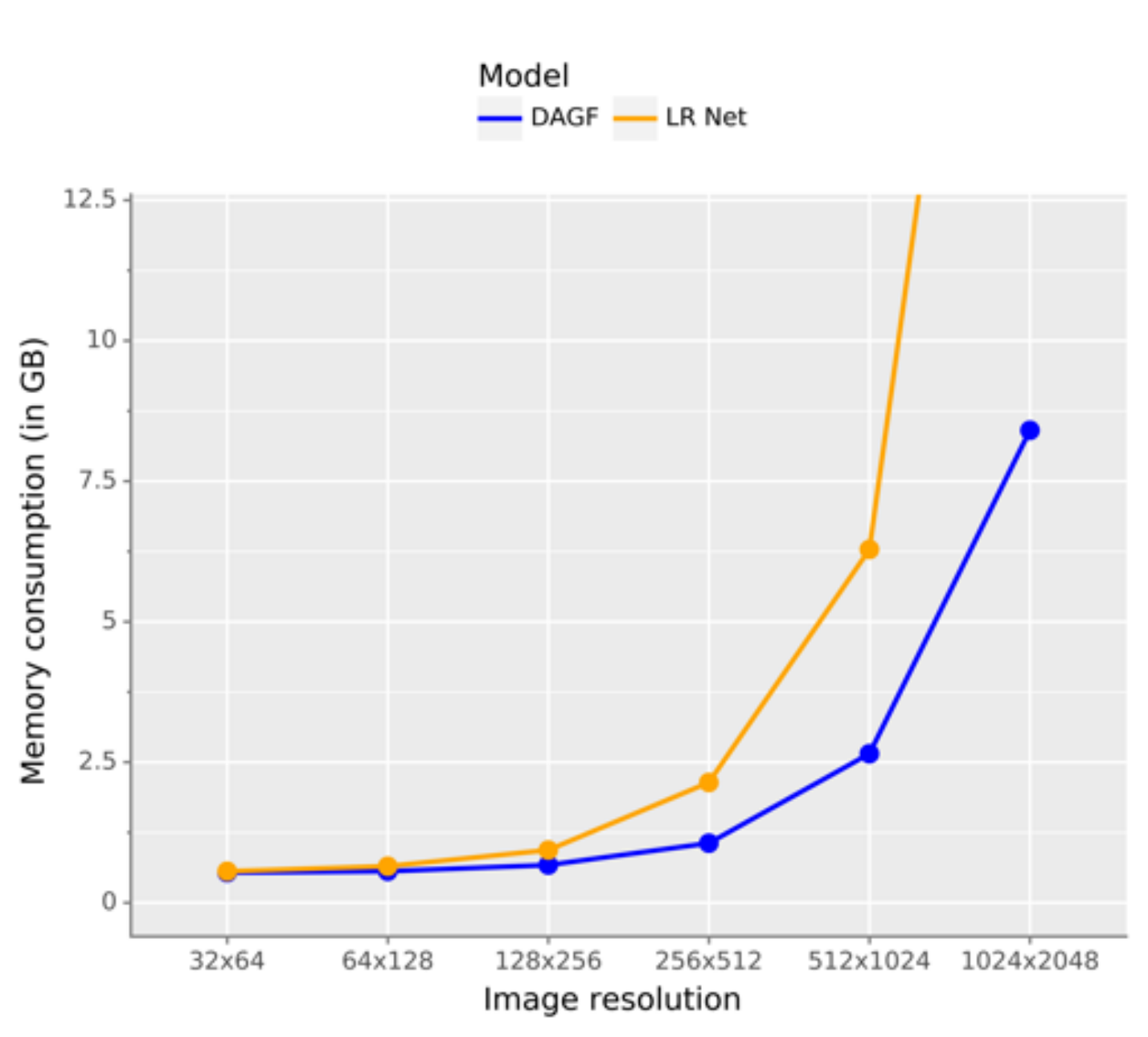}
    \caption{\textbf{Memory Consumption vs Image Size.} Without a guided filter backbone, LRNet does not scale to larger image sizes.}
    \label{fig:memory_consumption}
  \end{minipage}
  \hfill
  \begin{minipage}[b]{0.48\textwidth}
    \includegraphics[width=\textwidth]{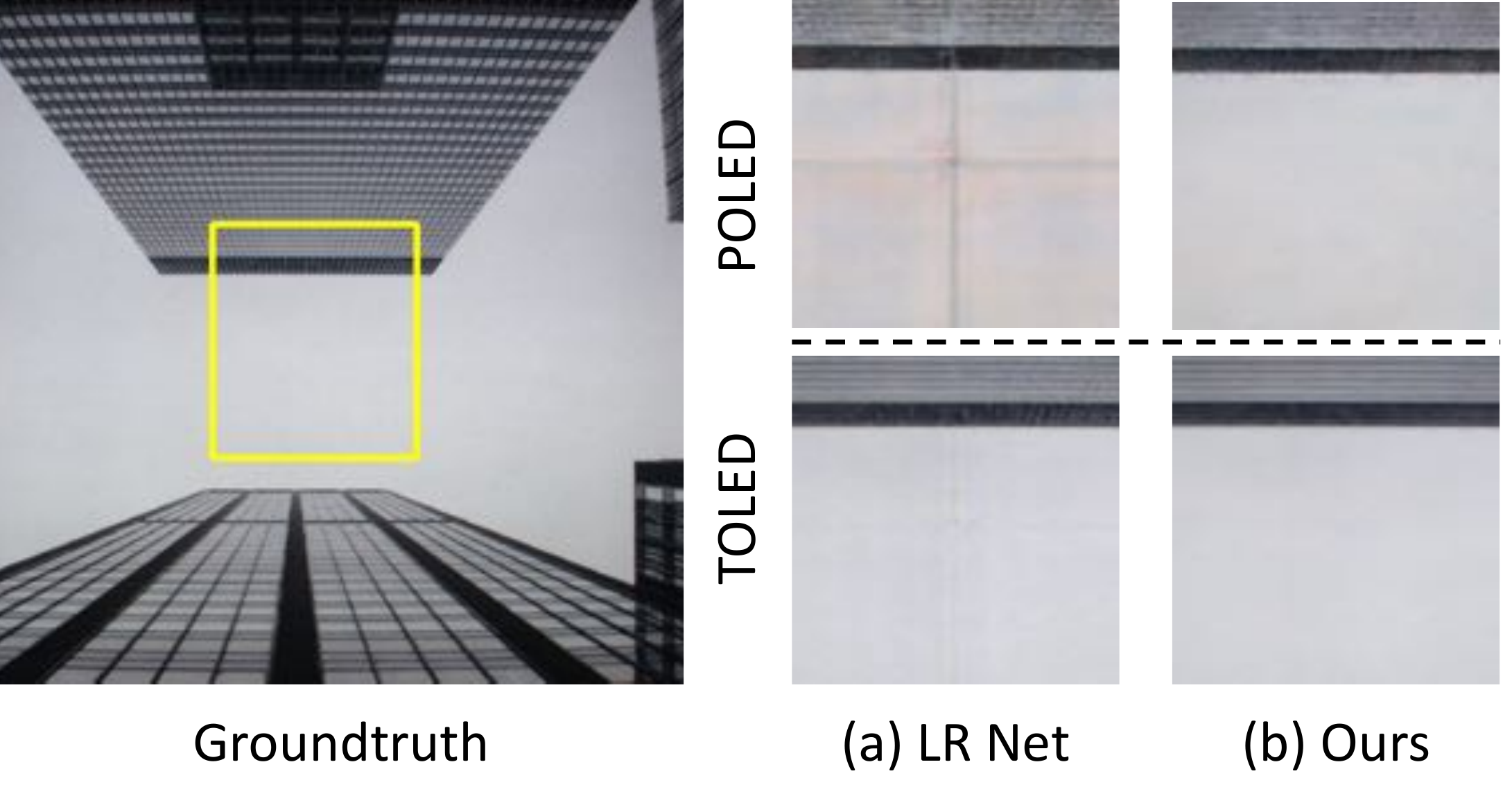}
    \caption{\textbf{Patch based methods lead to line artefacts, evident in the more challenging POLED output.} In contrast, our method operates on the entire image and produces no such artefacts.}
    \label{fig:patch_discont}
  \end{minipage}
\end{figure}

The guided filter allows our approach to directly use high resolution images as input, as opposed to operating on patches or downsampled inputs. To demonstrated its utility, we compare the performance obtained with and without a guided filter. Without a guided filter framework, LRNet must be trained patch-wise, due to memory constraints (Figure \ref{fig:memory_consumption}). At test time, we assemble the output patch-wise. 

Using a guided filter provides a significant benefit of $2.5$ dB and $1.8$ dB on POLED images and TOLED images, respectively (Table \ref{tab:guided_filter}). Although marginally better LPIPS metrics indicate that LRNet produces more visually pleasing outputs, line artefacts can be observed in the outputs (Figure \ref{fig:patch_discont}). Such artefacts are prominent in the more challenging POLED dataset. Alternative evaluation strategies for LRNet such as using overlapping patches followed by averaging, or feeding the entire high-resolution input results in blurry outputs and degrades performance. 

% We quantitatively show this in our supplementary material. Add once supplementary completed.

Table \ref{tab:guided_filter} also features comparisons against other transformation functions $F(X)$. Experiments indicate a clear advantage in using our atrous residual block over either $1 \times 1$ or $3 \times 3$ conv layers proposed in Wu \textit{et al.} \cite{DGF_wu2017fast}.

\begin{table}[!t]
\centering
\caption{Using a \textbf{trainable guided filter} provides greater context by scaling to larger image dimensions.}

\resizebox{0.95\textwidth}{!}{\begin{tabular}{ c||c c c|c c c }
\hline
\textbf{Backbone}& 
\multicolumn{3}{c}{\textbf{POLED}} & \multicolumn{3}{c}{\textbf{TOLED}}\\
\hline
{} & 
PSNR $\uparrow$ & SSIM $\uparrow$ & LPIPS $\downarrow$ & 
PSNR $\uparrow$ & SSIM $\uparrow$ & LPIPS $\downarrow$ \\
\hline

No Guided Filter & 
{30.14} & {0.938} & \textbf{0.212} & 
{33.92} & {0.963} & \textbf{0.132} \\
\hline

Conv 1x1 Guided Filter &
{32.39} & {0.940} & {0.223} &
{35.605} & {0.963} & {0.141} \\
\hline

Conv 3x3 Guided Filter & 
{32.50} & {0.942} & {0.220} & 
{35.84} & {0.965} & {0.150} \\ 
\hline

\textbf{Smoothed Atrous Block} &
\textbf{32.87} & \textbf{0.946} & {0.216} &
\textbf{35.87} & \textbf{0.966} & {0.147} \\
\hline
\end{tabular}}

\label{tab:guided_filter}
\end{table}

\subsection{Other Ablative Studies}\label{ablative_studies}

All ablative results are presented in Table \ref{tab:ablative_studies}.\\

\noindent\textbf{Smoothed Dilated Convolutions.} Using either $3 \times 3$ convolutions or exponentially growing dilation rates \cite{GCANet_chen2018gated,CAN_Chen_2017_ICCV} with the same number blocks leads to inferior performance. In contrast, parallel atrous convolutions lead to a larger receptive field at similar depth \cite{Brehm_2020_CVPR_Workshops} and improves performance. We also verify that introducing a smoothing operation before atrous convolutions is beneficial, and qualitatively leads to fewer gridding (or checkerboard) artefacts \cite{gridding_wacv_2018,gridding_wacv_2018_wang,griddingwang2018smoothed}.

\noindent\textbf{Residual and Gated Connections.} Consistent with Zhang \textit{et al.} \cite{RDN_zhang2020residual}, removing local residual connections leads to a considerable degradation in performance. Similarly, gated attention, which can be perceived as a global residual connection with tunable weights, provides a noticeable performance gain.

\noindent\textbf{Adaptive Normalisation.} Using adaptive equivalents of batch-norm \cite{ioffe2015batch} or instance-norm \cite{instance_ulyanov2017improved} improves performance. Experiments indicate a marginal increase of adaptive batch-norm over adaptive instance-norm. However, since we use smaller minibatch sizes while training on $1024 \times 2048$, we prefer adaptive instance-norm. 

\noindent\textbf{Channel attention.} Unlike recent variants of channel attention \cite{SENet_Hu_2018_CVPR,A2ANet_NIPS2018_7318,ECA_Wang_2020_CVPR,CBAM_Woo_2018_ECCV}, our implementation learns channel-wise weights, but does not capture inter-channel dependency. Experimenting with Efficient Channel Attention (ECA) \cite{ECA_Wang_2020_CVPR} did not confer any substantial benefit, indicating that modelling inter-channel dependencies may not be important in our problem. 

\noindent\textbf{Loss functions.} Compared to L1 loss, optimising MS-SSIM \cite{MS-SSIM_zhao2016loss} loss can improve PSNR marginally, but tends to be unstable during the early stages of training. Perceptual \cite{percep_adv_degrade_blau2018} and adversarial \cite{GANs_NIPS2014_5423} losses improve visual quality, reflected in better LPIPS scores, but degrade PSNR and SSIM metric performance \cite{percep_adv_degrade_blau2018}. Overall, L1 loss is a simple yet superior choice.

\begin{table}[!ht]
\centering
\caption{\textbf{Ablative Studies.}  We experiment with various components present in our approach to justify our architecture choices.}

    \resizebox{0.95\textwidth}{!}{\begin{tabular}{ c c c c c c||c c c|c c c }
    \hline
    \multicolumn{6}{c||}{\textbf{Conditions}}& 
    \multicolumn{3}{c|}{\textbf{POLED}} & 
    \multicolumn{3}{c}{\textbf{TOLED}}\\
    \hline
    \multicolumn{12}{c}{Smooth Atrous Convolutions} \\
    \hline
    
    \multicolumn{1}{m{1cm}}{\centering Atrous} & \multicolumn{2}{m{1.5cm}}{\centering Parallel Atrous} & \multicolumn{3}{m{1.8cm}||}{\centering Smooth Atrous} &
    PSNR $\uparrow$ & SSIM $\uparrow$ & LPIPS $\downarrow$ & 
    PSNR $\uparrow$ & SSIM $\uparrow$ & LPIPS $\downarrow$ \\
    \hline
    - & \multicolumn{2}{c}{-} & \multicolumn{3}{c||}{-} & 
    31.26 & 0.936 & 0.257 & 
    34.76 & 0.960 & 0.157\\
    $\checkmark$ & \multicolumn{2}{c}{-} & \multicolumn{3}{c||}{-} & 
    31.11 & 0.928 & 0.311 & 
    32.78 & 0.945 & 0.240\\
    $\checkmark$ & \multicolumn{2}{c}{$\checkmark$} & \multicolumn{3}{c||}{-} &
    32.39 & 0.943 & 0.233 & 
    35.46 & 0.963 & 0.157 \\
    $\checkmark$ & \multicolumn{2}{c}{$\checkmark$} & \multicolumn{3}{c||}{$\checkmark$} &
    \textbf{32.87} & \textbf{0.946} & \textbf{0.216} &
        \textbf{35.87} & \textbf{0.966} & \textbf{0.147}\\
    \hline
    
    \multicolumn{12}{c}{Residual and Gated Connections} \\
    \hline
    \multicolumn{3}{c}{Residual} & \multicolumn{3}{c||}{Gated} &
    PSNR $\uparrow$ & SSIM $\uparrow$ & LPIPS $\downarrow$ & 
    PSNR $\uparrow$ & SSIM $\uparrow$ & LPIPS $\downarrow$ \\
    \hline
    \multicolumn{3}{c}{-} & \multicolumn{3}{c||}{-} & 
    29.59 & 0.907 & 0.382 & 
    32.03 & 0.938 & 0.267 \\
    \multicolumn{3}{c}{$\checkmark$} & \multicolumn{3}{c||}{-} &
    32.19 & 0.941 & 0.255 & 
    35.14 & 0.961 & 0.162 \\
    \multicolumn{3}{c}{$\checkmark$} & \multicolumn{3}{c||}{$\checkmark$} &
    \textbf{32.87} & \textbf{0.946} & \textbf{0.216} &
    \textbf{35.87} & \textbf{0.966} & \textbf{0.147} \\
    \hline
    
    \multicolumn{12}{c}{Normalization Layers} \\
    \hline
    BN \cite{ioffe2015batch} & IN \cite{instance_ulyanov2017improved} & \multicolumn{2}{c}{ABN \cite{CAN_Chen_2017_ICCV}} & \multicolumn{2}{c||}{AIN} &
    PSNR $\uparrow$ & SSIM $\uparrow$ & LPIPS $\downarrow$ & 
    PSNR $\uparrow$ & SSIM $\uparrow$ & LPIPS $\downarrow$ \\
    \hline
    {-} & {-} & \multicolumn{2}{c}{-} & \multicolumn{2}{c||}{-} &
    {31.78} & {0.937} & {0.278} & 
    {35.09} & {0.96} & {0.165} \\
    {$\checkmark$} & {-} & \multicolumn{2}{c}{-} & \multicolumn{2}{c||}{-} &
    {30.75} & {0.919} & {0.268} &
    {33.20} &{0.943} & {0.191}\\
    {-} & {$\checkmark$} & \multicolumn{2}{c}{-} & \multicolumn{2}{c||}{-} &
    30.17 & 0.918 & 0.289 &
    30.54 & 0.92 & 0.224 \\
    {-} & {-} & \multicolumn{2}{c}{$\checkmark$} & \multicolumn{2}{c||}{-} &
    {32.73} & {0.945} & {0.225} &
    {\textbf{36.02}} & {\textbf{0.966}} & {\textbf{0.14}} \\
    {-} & {-} & \multicolumn{2}{c}{-} & \multicolumn{2}{c||}{$\checkmark$} &
    {\textbf{32.87}} & {\textbf{0.946}} & {\textbf{0.216}} &
    {35.87} & {\textbf{0.966}} & {0.147} \\
    \hline
    
    \multicolumn{12}{c}{Channel Attention} \\
    \hline
    \multicolumn{3}{c}{ECA \cite{ECA_Wang_2020_CVPR}} & \multicolumn{3}{c||}{FFA \cite{GCANet_chen2018gated}} &
    PSNR $\uparrow$ & SSIM $\uparrow$ & LPIPS $\downarrow$ & 
    PSNR $\uparrow$ & SSIM $\uparrow$ & LPIPS $\downarrow$ \\
    \hline 
    \multicolumn{3}{c}{-} & \multicolumn{3}{c||}{-} & 
    {32.62} & {0.944} & {0.225} & 
    {35.72} & {0.964} & {0.152} \\
    \multicolumn{3}{c}{$\checkmark$} & \multicolumn{3}{c||}{-} &
    {32.66} & {0.944} & {0.232} & 
    \textbf{35.98} & \textbf{0.966} & \textbf{0.143} \\
    \multicolumn{3}{c}{-} & \multicolumn{3}{c||}{$\checkmark$} &
    {\textbf{32.87}} & {\textbf{0.946}} & {\textbf{0.216}} &
    {35.87} & \textbf{0.966} & {0.147} \\
    \hline
    \multicolumn{12}{c}{Loss Functions} \\
    \hline
    {L1} & \multicolumn{1}{m{1.4cm}}{\centering MS-SSIM \cite{MS-SSIM_zhao2016loss}} & \multicolumn{2}{m{1.1cm}}{\centering Percep. \cite{johnson2016perceptual}} & \multicolumn{2}{m{1.1cm}||}{\centering Adv. \cite{GANs_NIPS2014_5423}} &
    PSNR $\uparrow$ & SSIM $\uparrow$ & LPIPS $\downarrow$ & 
    PSNR $\uparrow$ & SSIM $\uparrow$ & LPIPS $\downarrow$ \\
    \hline
    $\checkmark$ & {-} & \multicolumn{2}{c}{-} & \multicolumn{2}{c||}{-} &
    \textbf{32.87} & \textbf{0.946} & 0.216 &
    35.87 & 0.966 & 0.147 \\
    $\checkmark$ & $\checkmark$ & \multicolumn{2}{c}{-} & \multicolumn{2}{c||}{-} &
    {32.55} & {\textbf{0.946}} & {0.208} &
    {\textbf{36.20}} & {\textbf{0.968}} & {0.125} \\
    {$\checkmark$} & {-} & \multicolumn{2}{c}{$\checkmark$} & \multicolumn{2}{c||}{-} &
    {31.75} & {0.936} & {0.189} &
    {35.45} & {0.963} & {0.112} \\
    {$\checkmark$} & {-} & \multicolumn{2}{c}{$\checkmark$} & \multicolumn{2}{c||}{$\checkmark$} &
    {31.81} & {0.94} & \textbf{0.178} &
    {34.59} & {0.922} & {\textbf{0.086}} \\
    \hline
    \end{tabular}}
    
\label{tab:ablative_studies}
\end{table}

\section{Conclusions}\label{conclusions}

In this paper, we introduce a novel architecture for image restoration in Under Display Cameras. Deviating from existing patch-based image restoration methods, we show that there is a significant benefit in directly training on megapixel images. Incorporated in an end-to-end manner, a guided filter framework alleviates artefacts associated with patch based methods. We also show that a carefully designed low-resolution network utilising smoothed atrous convolutions and various attention blocks is essential for superior performance. Finally, we develop a simple simulation scheme to pre-train our model and boost performance. Our overall approach outperforms current models and attains 2nd place in the UDC 2020 Challenge- Track 2:POLED.  

As evidenced by our superlative performance on POLED restoration, the proposed method is more suited for higher degree of image degradation. Future work could address modifications to better handle a variety of image degradation tasks. Another promising perspective is to make better use of simulated data, for instance, in a domain-adaptation framework.

\noindent\textbf{Acknowledgements.} The authors would like to thank Genesis Cloud for providing additional compute hours.  

% ---- Bibliography ----
% INITIAL SUBMISSION 
% BibTeX users should specify bibliography style 'splncs04'.
% References will then be sorted and formatted in the correct style.
% Comment following two lines in the camera-ready version
% \bibliographystyle{toolkit_includes/splncs04}
% \bibliography{ref}

% CAMERA READY - Use bbl instead of bib for camera-ready
% Uncomment

\end{document}

% \renewcommand\thelinenumber{\color[rgb]{0.2,0.5,0.8}\normalfont\sffamily\scriptsize\arabic{linenumber}\color[rgb]{0,0,0}}
% \renewcommand\makeLineNumber {\hss\thelinenumber\ \hspace{6mm} \rlap{\hskip\textwidth\ \hspace{6.5mm}\thelinenumber}}
% \linenumbers
\pagestyle{headings}
\mainmatter
\def\ECCVSubNumber{32}  % Insert your submission number here

\title{[Supplementary Material]\\
Deep Atrous Guided Filter for Image Restoration in Under Display Cameras} % Replace with your title

%******************
% INITIAL SUBMISSION 
\begin{comment}
\titlerunning{ECCV-20 submission ID \ECCVSubNumber} 
\authorrunning{ECCV-20 submission ID \ECCVSubNumber} 
\author{Anonymous ECCV submission}
\institute{Paper ID \ECCVSubNumber}
\end{comment}

%******************
% CAMERA READY SUBMISSION
% TODO: Add author names post acceptance
% \begin{comment}
\titlerunning{Deep Atrous Guided Filter}
% If the paper title is too long for the running head, you can set
% an abbreviated paper title here
%
\author{Varun Sundar\thanks{Equal Contribution} \and
Sumanth Hegde\printfnsymbol{1} \and
Divya Kothandaraman \and
Kaushik Mitra}
\authorrunning{V. Sundar et al.}
% First names are abbreviated in the running head.
% If there are more than two authors, 'et al.' is used.

\institute{Indian Institute of Technology Madras \\
\email{\{varunsundar@smail, sumanth@smail, ee15b085@smail, kmitra@ee\}.iitm.ac.in}\\
\url{https://github.com/varun19299/deep-atrous-guided-filter}}
% \email{\{varunsundar,sumanth, divya\_kothandaraman\}@smail.iitm.ac.in, kmitra@ee.iitm.ac.in}}
% \end{comment}

%******************
\maketitle
\section{Guided Filter Details}
In this section, we present the algorithmic details of our trainable guided filter, which uses the high-resolution input as the guide image and the restored low-resolution output as the filtering input, to produce the high-resolution output via joint-upsampling. \\

\noindent \textbf{Architecture details:} We first detail the architectural choices made for the trainable components of the guided filter. Our implementation is similar to He \textit{et al.} \cite{he2015fast}, except that the mean filter $f_\mu$ is implemented via a 3x3 convolutional layer and the transformation function $F(.)$ via our atrous residual block. Finally, the local parameter estimator $f_{\text{local}}$ consists of a 3 layer, $1 \times 1$ convolutional block, with adaptive normalisation layers and ReLU activations in between. Similar to Wu \textit{et al.} \cite{DGF_wu2017fast}, the guided filter is trained in an end-to-end manner with LRNet. The complete architecture of $f_{\text{local}}$ is detailed in Table \ref{supp_tab:f_local}.

\begin{table}
    \centering
    \caption{\textbf{Architecture of $f_{\text{local}}$.}}
    
    \resizebox{0.8\textwidth}{!}{\begin{tabular}{c||c|c|c|c}
    \hline
      Layer &  Convolution & Adaptive Norm & ReLU & \multicolumn{1}{m{3cm}}{\centering Input / Output Channel Size}\\
    \hline
    1 & 1x1 & \checkmark & \checkmark & 3+3 / 32\\
    2 & 1x1 &  \checkmark & \checkmark & 32 / 32\\
    3 & 1x1 & - & - & 32 / 3\\
    \hline
    \end{tabular}}
    \label{supp_tab:f_local}
\end{table}
% The algorithm proceeds in five key steps as follows: {\red should we briefly explain the algorithm in words?}
% i don't think it's reqd

\noindent\textbf{Algorithm of the Guided Filter Network:} The entire algorithm is outlined in Algorithm \ref{alg:trainable_gf}. Here, $f_\uparrow$ denotes an upsampling operation (we use bilinear upsampling). $[\cdot, \cdot]$ denotes concatenation and $\odot$ denotes the Hadamard product. \\

\begin{algorithm}[t]
\SetKwInOut{Input}{Input}  
\SetKwInOut{Notation}{Notation}
\Notation{Learnt Mean Filter $f_\mu$\\
Transformation Function $F$\\
Local Parameter Estimator $f_\text{local}$\\ 
Bilinear Upsampling $f_\uparrow$  \\
Concatenation operation $[\cdot, \cdot]$
}
\Input{Low-resolution Image $X_l$\\
High resolution Image $X_h$\\
Low-resolution Output $Y_l$  
}

\setstretch{1.35}
\SetKwInOut{Output}{Output}
\Output{High-resolution Output $Y_h$}
 $G_l = F(X_l)$ , $G_h = F(X_h)$\\
 $\overline{G_l} = f_\mu (G_l)$ \\
\nonl $\overline{Y_l} = f_\mu (Y_l)$ \\
\nonl $\overline{G^{2}_l} = f_\mu (G_l \odot G_l)$ \\
\nonl $ \overline{G_l Y_l} = f_\mu (G_l \odot Y_l)$ \\
 $\Sigma_{G_lG_l} = \overline{G^{2}_l} - \overline{G_l} \odot \overline{G_l}$\\
\nonl $\Sigma_{G_l Y_l} = \overline{G_l Y_l} - \overline{G_l} \odot \overline{Y_l}$\\
 $A_l = f_{\text{local}}([\Sigma_{G_lG_l}, \Sigma_{G_l Y_l}])$\\
 \nonl $b_l = \overline{Y_l} - A_l \odot \overline{G_l} $\\
 $A_h = f_{\uparrow} (A_l) ~,~ b_h = f_{\uparrow} (b_l)$
 
 $Y_h = A_h \odot G_h + b_h$
\caption{Trainable Guided Filter Network}
\label{alg:trainable_gf}
\end{algorithm}
\section{Line Artefacts in Patch-Based Methods}\label{supp:line_artefacts}

\begin{figure*}[!t]
    \centering
    \includegraphics[width=\textwidth]{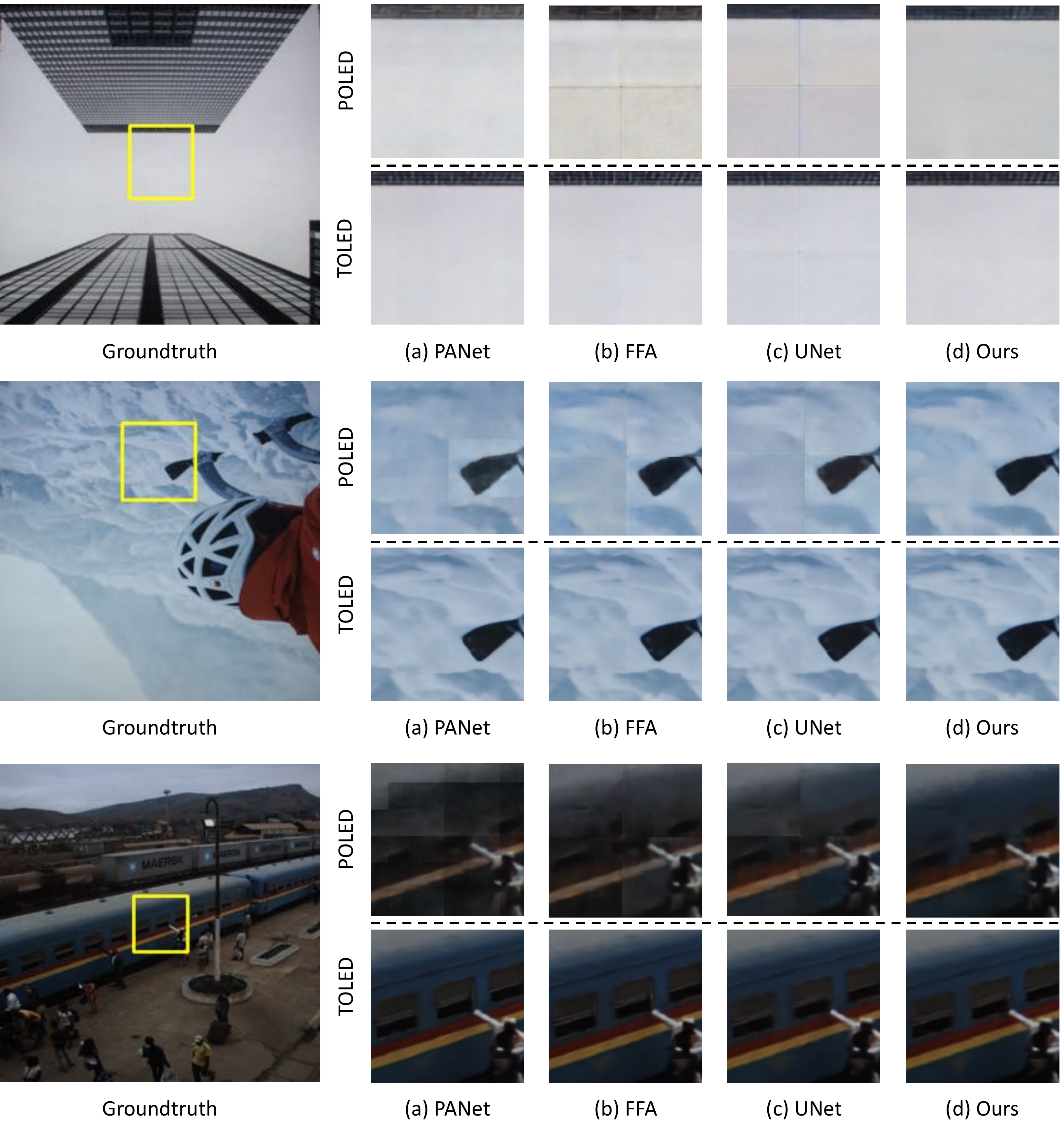}
    \caption{\textbf{Line artefacts shown for patch-based baseline methods.} Our proposed method lacks such artefacts, since it directly trains on the entire megapixel input.}
    \label{supp_fig:patch_comp}
\end{figure*}

Patch-based methods lead to line artefacts, especially evident in severe degradation scenarios. We attribute this to the limited context available to patch-based methods during training. Our approach plays a significant role in alleviating these artefacts. Expanding on the comparisons shown in Section 5.1 of the main paper, we show similar comparisons against the other patch-based baseline methods, viz. PANet \cite{PANet_mei2020pyramid}, FFA-Net \cite{qin2020ffa} and UNet \cite{UNet_ronneberger2015u} (Figure \ref{supp_fig:patch_comp}).

\section{Simulation Dataset}

\begin{figure*}[!t]
    \centering
    \includegraphics[width=\textwidth]{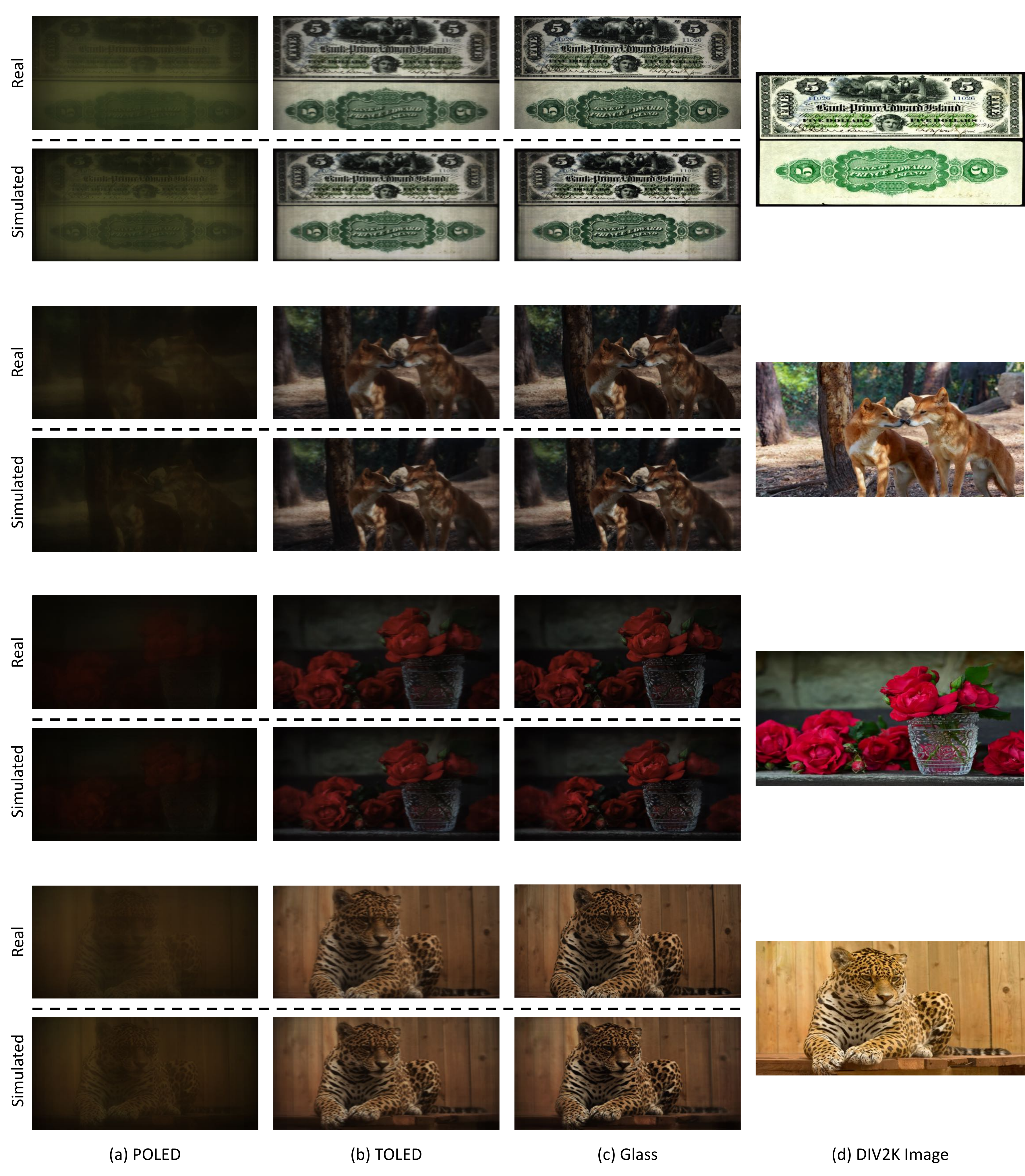}
    \caption{\textbf{More Simulation Outputs.}}
    \label{supp_fig:sim_comp}
\end{figure*}

We show more outputs from our simulation procedure in Figure \ref{supp_fig:sim_comp}, comparing it against corresponding real measurements. We can observe that our simulated outputs are perceptually similar to real measurements, and also bear similar artefacts such as low-light degradation in POLED and stripe bands in TOLED. Notice that while the simulated measurements align with the clean DIV2K \cite{DIV2K_Agustsson_2017_CVPR_Workshops} images, the real measurements do not. 

% ---- Bibliography ----
% INITIAL SUBMISSION 
% BibTeX users should specify bibliography style 'splncs04'.
% References will then be sorted and formatted in the correct style.
% Comment following two lines in the camera-ready version
% \bibliographystyle{toolkit_includes/splncs04}
% \bibliography{ref}

% CAMERA READY - Use bbl instead of bib for camera-ready
% Uncomment